\documentclass[conference]{IEEEtran}
\IEEEoverridecommandlockouts
\usepackage{cite}
\usepackage{amsmath,amssymb}
\usepackage{graphicx}
\usepackage{booktabs}
\usepackage{textcomp}
\usepackage{xcolor}
\usepackage[hidelinks]{hyperref}
\hypersetup{pdfauthor={},pdfsubject={},pdfkeywords={},pdfcreator={}}
\graphicspath{{../results/}{./}}
\newcommand{\NullB}{200}
\newcommand{\BootR}{200}
\newcommand{\NItems}{4{,}000}
\newcommand{\PrimaryK}{92}
\newcommand{\NGlomFull}{51}
\newcommand{\NCellsFull}{1{,}886}

\newcommand{\FanMax}{473}
\newcommand{\FanMaxGlom}{DC1}
\newcommand{\NOdorants}{172}
\newcommand{\MissingPct}{28.7\%}
\newcommand{\DoorCommit}{db323a4}
\newcommand{\PrimaryReal}{0.4063}
\newcommand{\PrimaryNullMean}{0.4148}
\newcommand{\PrimaryNullSD}{0.0035}

\newcommand{\PrimaryRel}{-2.06\%}

\newcommand{\PrimaryPtwo}{0.020}

\newcommand{\BootLo}{-3.1\%}
\newcommand{\BootHi}{-0.8\%}

\newcommand{\MarginPrimary}{5\%}
\newcommand{\MdeUp}{3.0\%}
\newcommand{\MdeDown}{2.5\%}

\newcommand{\PrimaryPHolmAll}{0.100}

\newcommand{\PlateauSweeps}{1}

\newcommand{\CovNew}{100\%}

\newcommand{\CovDatasets}{100}

\newcommand{\ImputeRelMin}{-0.7\%}
\newcommand{\ImputeRelMax}{+0.6\%}
\newcommand{\ImputePMin}{0.439}
\newcommand{\RobN}{26}
\newcommand{\RobB}{40}
\newcommand{\RobRelMin}{-3.9\%}
\newcommand{\RobRelMax}{+0.8\%}

\newcommand{\ArchKbitsRatio}{1.37}
\newcommand{\ArchStorageRatio}{0.90}
\newcommand{\ArchComputeRatio}{1.33}

\newcommand{\WiringRows}{8 & 0.1973 & 0.2034 $\pm$ 0.0085 & -3.0 & 0.468 & 1.000 & +3.7 & +26.7 & +3.2 \\
16 & 0.2879 & 0.2933 $\pm$ 0.0062 & -1.8 & 0.438 & 1.000 & +2.7 & +9.7 & +3.2 \\
32 & 0.3531 & 0.3575 $\pm$ 0.0045 & -1.2 & 0.338 & 1.000 & +3.3 & +3.0 & +3.6 \\
64 & 0.3963 & 0.4001 $\pm$ 0.0039 & -0.9 & 0.348 & 1.000 & +3.4 & +0.7 & +3.4 \\
92$^\ast$ & 0.4063 & 0.4148 $\pm$ 0.0035 & -2.1 & 0.020 & -- & +3.2 & +0.1 & +3.2 \\}
\newcommand{\ArchRows}{8 & 0.197 & 0.019 & 0.264 (72) & 0.305 (102) \\
16 & 0.288 & 0.069 & 0.340 (130) & 0.305 (102) \\
32 & 0.353 & 0.151 & 0.394 (229) & 0.305 (102) \\
64 & 0.396 & 0.249 & 0.436 (397) & 0.305 (102) \\
92 & 0.406 & 0.296 & 0.454 (523) & 0.305 (102) \\}
\newcommand{\RobRows}{Baseline & -2.2 & 0.049 \\
Left hemisphere & -1.1 & 0.244 \\
$\geq$5-synapse threshold & -2.3 & 0.098 \\
Missing data (3 imputations) & -0.7 to +0.6 & 0.439--0.780 \\
Coverage thresholds (2) & -1.7 to -1.4 & 0.049--0.341 \\
All 46 glomeruli with data, imputed & -0.0 & 1.000 \\
Mixture seeds (4) & -2.4 to -1.6 & 0.049--0.244 \\
Mixture size 2 / 2--10 & -2.0 to -0.9 & 0.146--0.195 \\
Measured odorants only & -3.9 & 0.049 \\
Normalised / angular metric & -2.4 to -1.7 & 0.049 \\
Tie rules (index, include, 2 seeds) & -2.4 to -2.1 & 0.049 \\
Synapse-weighted & -1.1 & 0.488 \\
Synthetic inputs (3) & -3.1 to +0.8 & 0.049--0.244 \\}

\newcommand{\ReplMnistRatio}{3.1}

\newcommand{\ReplMnistFlyRetr}{0.365}
\newcommand{\ReplMnistRatioCentred}{3.1}
\newcommand{\ReplMnistRecallRatio}{1.7}

\newcommand{\ReplMnistLshRetr}{0.159}

\newcommand{\ReplOdourRatio}{1.7}
\newcommand{\ReplTrials}{50}
\newcommand{\ReplOpsMin}{0.31}
\newcommand{\ReplOpsMax}{0.81}

\newcommand{\ConnB}{50}

\newcommand{\ConnOpsBits}{98}
\newcommand{\ConnOpsWins}{4}
\newcommand{\ConnNData}{4}

\newcommand{\ConnOdourOpsBits}{102}

\newcommand{\IdxRatioEight}{0.31}
\newcommand{\IdxRatioSixteen}{1.01}
\newcommand{\TieRatioEight}{0.07}
\newcommand{\TieRatioSixteen}{1.04}

\newcommand{\NoCentSmallMin}{-7\%}
\newcommand{\NoCentSmallMax}{-5\%}
\newcommand{\NoCentLargeAbs}{2\%}

\newcommand{\AnimalRows}{MaleCNS R & m & 1{,}886 & 5.28 & 0.53 & +2.7 & -2.2 (0.015) & -2.6 (0.020) & -1.3 & 2.3 \\
MaleCNS L & m & 1{,}865 & 5.30 & 0.54 & +4.8 & -0.9 (0.294) & -0.8 (0.376) & -1.0 & 2.2 \\
Hemibrain R & f & 1{,}761 & 6.05 & 0.50 & +0.0 & -0.7 (0.491) & -0.0 (0.990) & -1.1 & 2.3 \\
FlyWire R & f & 2{,}406 & 5.00 & 0.47 & +11.3 & -2.7 (0.001) & -2.7 (0.020) & -0.7 & 2.4 \\
FlyWire L & f & 2{,}378 & 5.12 & 0.49 & +17.8 & -1.6 (0.037) & -1.6 (0.079) & -2.0 & 2.9 \\
BANC R & f & 2{,}046 & 5.03 & 0.57 & +6.2 & -3.1 (0.003) & -3.0 (0.020) & -3.2 & 2.3 \\
BANC L & f & 2{,}089 & 5.18 & 0.55 & +8.7 & -1.3 (0.168) & -2.2 (0.059) & -3.1 & 2.3 \\}
\newcommand{\AnimOdourHolmMin}{0.007}
\newcommand{\AnimMatchedHolmMin}{0.139}

\newcommand{\AnimOdourHolmN}{2}
\newcommand{\AnimOdourHolmNames}{FlyWire R and BANC R}
\newcommand{\AnimQHolmMax}{0.010}
\newcommand{\HemiQzTwo}{+2.9}
\newcommand{\HemiQpTwo}{0.002}
\newcommand{\QDepartTwo}{7}
\newcommand{\AnimN}{7}
\newcommand{\AnimAnimals}{4}
\newcommand{\AnimB}{100}
\newcommand{\AnimBBench}{50}
\newcommand{\AnimBQ}{1000}
\newcommand{\AnimBOdour}{2000}
\newcommand{\AnimTrials}{5}

\newcommand{\AnimOdourPFloor}{0.001}
\newcommand{\AnimQNSigHolm}{6}
\newcommand{\AnimCommon}{49}
\newcommand{\AnimCommonOdour}{34}

\newcommand{\AnimNNeg}{134}
\newcommand{\AnimRelMedian}{-1.6\%}
\newcommand{\AnimRelIQR}{-3.2\% to -0.3\%}
\newcommand{\AnimNProto}{168}
\newcommand{\AnimNFloor}{59}
\newcommand{\AnimNFloorUp}{2}
\newcommand{\AnimOdourMin}{-3.1\%}
\newcommand{\AnimOdourMax}{-0.7\%}

\newcommand{\AnimOdourNSig}{4}
\newcommand{\AnimMatchedMin}{-3.0\%}
\newcommand{\AnimMatchedMax}{-0.0\%}

\newcommand{\AnimalsProtoNeg}{4}
\newcommand{\AnimQzMin}{0.0}
\newcommand{\AnimQzMax}{17.8}
\newcommand{\AnimQNSig}{6}

\newcommand{\AnimLshMin}{2.2}
\newcommand{\AnimLshMax}{2.9}
\newcommand{\AnimCellsMin}{1{,}761}
\newcommand{\AnimCellsMax}{2{,}406}
\newcommand{\AnimInMin}{5.00}
\newcommand{\AnimInMax}{6.05}
\newcommand{\CtrlB}{10}
\newcommand{\CtrlTrials}{20}
\newcommand{\CtrlOutMin}{+3\%}
\newcommand{\CtrlOutMax}{+17\%}
\newcommand{\CtrlOutPos}{17}

\newcommand{\CtrlOutOdMin}{+1\%}
\newcommand{\CtrlOutOdMax}{+3\%}
\newcommand{\CtrlOutOdPos}{4}
\newcommand{\CtrlInMin}{-7\%}
\newcommand{\CtrlInMax}{+8\%}
\newcommand{\CtrlInPos}{2}
\newcommand{\CtrlInNeg}{11}
\newcommand{\CtrlInOdMin}{-3\%}
\newcommand{\CtrlInOdMax}{+19\%}

\newcommand{\CtrlCells}{18}
\newcommand{\CtrlOdCells}{6}

\newcommand{\AnimOutLo}{+3.9\%}
\newcommand{\AnimOutHi}{+6.4\%}

\newcommand{\AnimInLo}{-4.2\%}
\newcommand{\AnimInHi}{-1.1\%}

\newcommand{\AnimCtrlN}{7}
\newcommand{\FanCommon}{49}
\newcommand{\FanBetweenMin}{0.73}
\newcommand{\FanBetweenMax}{0.95}
\newcommand{\FanBetweenMed}{0.86}
\newcommand{\FanWithinMin}{0.80}
\newcommand{\FanWithinMax}{0.94}
\newcommand{\FanSpread}{17}
\newcommand{\FanTop}{DP1m, DM1, DC1, DM2, VA2}
\newcommand{\FanBottom}{DA4m, DA3, VL1}
\newcommand{\FanWeightPos}{6}
\newcommand{\FanN}{7}
\newcommand{\FanWeightSig}{2}
\newcommand{\FanWeightMin}{-0.13}
\newcommand{\FanWeightMax}{+0.66}

\newcommand{\FanBreadthMin}{+0.13}
\newcommand{\FanBreadthMax}{+0.30}
\newcommand{\FanSdMin}{+0.24}
\newcommand{\FanSdMax}{+0.40}
\newcommand{\FanSdSig}{2}

\newcommand{\FanBreadthPmin}{0.079}
\newcommand{\TaskNovConn}{-0.006 [-0.016, +0.003]}
\newcommand{\TaskNovOut}{+0.003 [-0.010, +0.017]}
\newcommand{\TaskNovIn}{-0.015 [-0.023, -0.008]}
\newcommand{\TaskNovNull}{0.741}
\newcommand{\TaskKnnConn}{-0.009 [-0.017, -0.000]}
\newcommand{\TaskKnnOut}{+0.010 [+0.004, +0.017]}
\newcommand{\TaskKnnIn}{-0.009 [-0.013, -0.003]}
\newcommand{\TaskKnnNull}{0.700}
\newcommand{\TaskOdNovConn}{-0.003 [-0.007, +0.001]}
\newcommand{\TaskOdNovOut}{-0.007 [-0.014, -0.001]}
\newcommand{\TaskOdNovIn}{-0.001 [-0.005, +0.003]}
\newcommand{\TaskOdNovNull}{0.560}
\newcommand{\TaskTrials}{10}
\newcommand{\TaskKMnist}{94}
\newcommand{\TaskKOdour}{92}
\newcommand{\TaskRows}{Novelty, MNIST (AUC) & 0.741 & -0.006 & +0.003 & -0.015$^\dagger$ & -0.012$^\dagger$ & -0.009 \\
FlyNN-style, MNIST (accuracy) & 0.700 & -0.009$^\dagger$ & +0.010$^\dagger$ & -0.009$^\dagger$ & -0.001 & +0.013$^\dagger$ \\
Novelty, odours (AUC) & 0.560 & -0.003 & -0.007$^\dagger$ & -0.001 & -0.007$^\dagger$ & -0.002 \\}

\begin{document}

\title{A Connectome Test of the Fly Hashing Algorithm}

\author{\IEEEauthorblockN{Sebastian Senge}
\IEEEauthorblockA{\textit{Independent Researcher}}}

\maketitle

\begin{abstract}
Dasgupta, Stevens and Navlakha (2017) showed that the \textit{Drosophila}
olfactory circuit, modelled as a random sparse projection followed by
winner-take-all, is a locality-sensitive hash that beats classical LSH. The
projection was random because the wiring was unknown. We test it against four
electron-microscopy connectomes (MaleCNS, hemibrain, FlyWire, BANC; four
animals, seven hemispheres) and report three findings. First, the 2017 pattern holds in a reimplementation of its protocol on SIFT, MNIST and odour mixtures (GloVe is near chance at short codes for every method): the fly hash beats $k$ Gaussian projections at short hash lengths
($\ReplMnistRatio\times$ in AP@200 on MNIST at $k=4$). Second, against the tested real-valued Gaussian baseline that advantage is per
active cell, not per operation: Gaussian projections given the same projection
arithmetic retrieve better on every dataset and at every input dimension
tested. Third, across four connectomes the measured pairing of glomeruli gives
no consistent retrieval advantage over degree-preserving rewiring: retrieval
is slightly lower (median \AnimRelMedian{}), and the small odour deficits
depend on how missing odour responses are treated. Separately, equalising
glomerular fan-out at fixed connection count improves retrieval in the model in
every hemisphere, while equalising inputs per cell lowers it on average. Yet the fan-out profile is similar across the four sampled animals (median between-animal $\rho=\FanBetweenMed$), structural synapse counts do not offset it, and its relation to odour tuning is weak. In the sampled animals the circuit thus allocates Kenyon cells unevenly in a similar way, and in the model that unevenness costs retrieval. For practice: in the tested connectomes, measured wiring gives no consistent retrieval advantage over degree-preserving random wiring, so a fly hash needs no connectome data, and its advantage is per active unit, which may suit hardware where active units rather than arithmetic are the binding cost, a hypothesis we do not test.
\end{abstract}


\section{Introduction}

\begin{figure*}[t]
\centerline{\includegraphics[width=\textwidth]{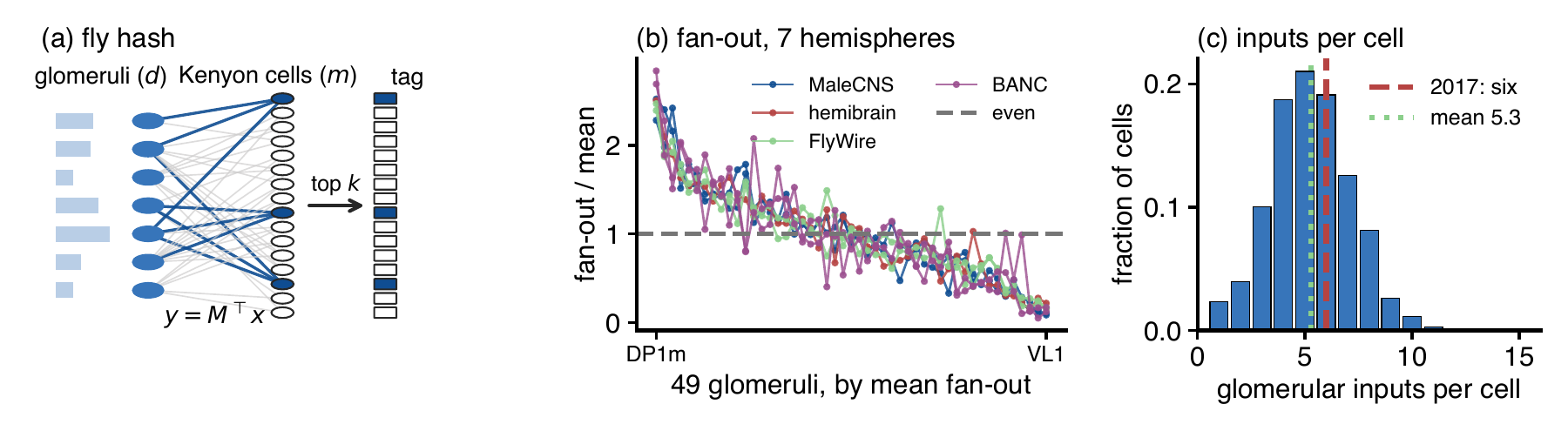}}
\caption{(a) The fly hash: glomerular input $x$ is expanded through a binary
projection $M$ onto Kenyon cells and the $k$ most active cells form the tag.
(b) Glomeruli reach very different numbers of Kenyon cells, and the profile is similar across the sampled hemispheres: relative fan-out of the $\FanCommon$ shared glomeruli in all seven, ordered by the mean; the dashed line is even fan-out.
(c) Inputs per Kenyon cell, against the six of the 2017 construction.}
\label{fig:concept}
\end{figure*}

Locality-sensitive hashing (LSH) maps items to short codes under which similar
items collide with high probability \cite{indyk1998,charikar2002}. Dasgupta
\textit{et al.} \cite{dasgupta2017} noticed that the fly olfactory circuit
computes a related transform (Fig.~\ref{fig:concept}a; Section~\ref{sec:bg}). On SIFT, GloVe and MNIST this fly hash
beat classical LSH by up to about $3\times$. The projection was random,
following the account that the wiring is effectively random \cite{caron2013}.

That account has since been revised: projection-neuron input to Kenyon cells is
structured \cite{zheng2022}, and four connectomes now give the complete
projection \cite{malecns2026,scheffer2020,dorkenwald2024,bates2026}. We ask
whether the random matrix was a harmless idealisation for the task that
motivated it; \cite{zheng2022} studied odour discrimination, not retrieval.
Our contributions:

\begin{itemize}
\item a reimplementation of the 2017 protocol with documented provenance,
extended with odour mixtures, that recovers the published pattern;
\item a budget analysis showing that, against real-valued Gaussian projections,
the fly hash's advantage is per active cell and not per operation;
\item a comparison of the fly hash across four connectomes and seven
hemispheres, on the 2017 benchmarks and on measured odour responses, against
degree-preserving rewiring and against controls that change one degree
sequence at a time at equal connection count;
\item the observation that the fan-out skew that costs retrieval in the model is similar across the four sampled animals, is not offset by structural synapse counts, and is only weakly related to odour tuning.
\end{itemize}

\section{Background}
\label{sec:bg}

\emph{Connectomes.} A connectome is a wiring diagram at the resolution of
individual neurons and synapses. It is reconstructed from electron-microscopy images by tracing every neuron and detecting every synapse. For the adult fruit fly this became feasible only recently; the four connectomes used here each contain the complete mushroom body.

\emph{The fly olfactory pathway.} An odour activates olfactory receptor neurons
in the antennae. Neurons that express the same receptor type all converge onto
one small, anatomically identifiable structure in the antennal lobe, a
\emph{glomerulus}; the fly has about $50$ olfactory glomeruli, so an odour is
first represented as a vector of about $50$ glomerular activities. Projection
neurons carry each glomerulus's signal to the mushroom body, where about
$2{,}000$ \emph{Kenyon cells} per hemisphere each receive input from a handful
of glomeruli. A single inhibitory neuron (APL) feeds back onto all Kenyon cells,
so that only the most strongly driven few percent fire. This sparse set of active Kenyon cells is the fly's code for the odour and the basis of associative memory \cite{dasgupta2017,caron2013}. In this paper $M$ is
the glomerulus-to-Kenyon-cell part of that pathway: which glomeruli each
Kenyon cell receives input from.

\emph{Similarity search and hashing.} Nearest-neighbour search asks, for a
query item, which stored items are most similar to it. Exhaustive comparison
is expensive for large collections, so approximate methods first map items to
short \emph{hashes} and compare those instead. A hash is locality-sensitive
if similar items receive similar hashes with high probability
\cite{indyk1998}. The classical construction multiplies the input by a random
Gaussian matrix and keeps the $k$ projected values, or their signs, as the hash
\cite{charikar2002}. The fly hash of \cite{dasgupta2017} replaces the dense
matrix with the sparse binary glomerulus-to-Kenyon-cell matrix, expands rather
than compresses the dimension, and keeps as the hash the identities of the $k$
most active Kenyon cells (Fig.~\ref{fig:concept}a). Retrieval quality is the
agreement between the neighbours found through the hashes and the true
neighbours.

\section{Related work}

The fly hash has been extended with dense pseudo-hashes and multi-probe
binning \cite{sharma2018}, and with projections learned by local plasticity
rules \cite{ryali2020,liang2021}; these change the algorithm, whereas we keep
it and change only the matrix. On the biological side, random convergence of
projection-neuron input was reported from anatomical tracing
\cite{caron2013}; electron microscopy later mapped the mushroom body
completely in the larva \cite{eichler2017} and in the adult hemibrain
\cite{li2020}, and found structured sampling in FAFB, where simulated odour
discrimination was worse than with random wiring \cite{zheng2022}. None of
these studies evaluated similarity search, the task behind the hashing
interpretation. Theory of expansion layers predicts that a small number of
inputs per cell is optimal for the mushroom body and the cerebellum
\cite{litwinkumar2017}; our controls instead vary how evenly inputs and
outputs are distributed at a fixed total.

\section{Methods}

\subsection{Connectomes and matrices}

Let $M\in\{0,1\}^{d\times m}$ with $M_{ji}=1$ iff glomerulus $j$ connects to
Kenyon cell $i$. For every connectome the same rule applies: uniglomerular
olfactory projection neurons are assigned to glomeruli by cell type
(multiglomerular, thermo- and hygrosensory types excluded), Kenyon cells are
those annotated on the same side, every reconstructed connection counts (no
synapse threshold), and cells without olfactory input are dropped. Hemibrain
glomerulus names are mapped to current nomenclature via the cross-dataset
matches of \cite{schlegel2024}. We use MaleCNS v1.0 (male, both hemispheres),
hemibrain v1.2 (female, right), FlyWire v783 and BANC v888 (female, both). All
share $\AnimCommon$ glomeruli; they have \AnimCellsMin{} to \AnimCellsMax{}
Kenyon cells with \AnimInMin{} to \AnimInMax{} inputs per cell. The datasets
differ in imaging, synapse detection and proofreading, so part of any
difference between them is technical. The primary connectome is MaleCNS right
($d=\NGlomFull$, $m=\NCellsFull$).

Only $M$ changes between conditions. \emph{The null} is uniform over binary
matrices with the same row and column sums as $M$, sampled by curveball trades
\cite{strona2014,carstens2015} (30 per cell; co-occurrence
$Q=\sum_{j<l}(MM^\top)_{jl}^2$ is stable from \PlateauSweeps{} sweep on); it
keeps both degree sequences and randomises which glomeruli share a cell, and
tests against it are approximate Monte Carlo randomization tests.
\emph{Equal-connection controls} have exactly $\mathrm{nnz}(M)$ ones with
inputs per cell made even, fan-out made even, or both, and random pairing.
\emph{The 2017 construction} gives each cell six uniform inputs, changing both
degree sequences and the connection count.

\subsection{Hash, scores and budgets}

The fly hash computes $y=M^\top x$ and sets the $k$ largest entries of $y$ to
one; cells tied at the threshold drive are taken in index order (a random
tie priority is a sensitivity analysis). Retrieval is scored by $\mathrm{AP@}n=\frac1n\sum_i
\mathrm{prec@}i\cdot\mathrm{rel}(i)$ over the $n$ true neighbours, so neighbours
never retrieved count as misses, together with recall@$n$. LSH computes $k$
Gaussian projections kept as real values, as in the 2017 comparison, or their
signs for bit codes. Budgets are per active cell ($k$ projections for $k$
winners), per stored bit, and per operation, counted as projection arithmetic
as in \cite{dasgupta2017}: $d$ multiplications and $d$ additions per
projection, one addition per non-zero of $M$; winner selection, retrieval and
any PCA fit are excluded.

\subsection{Approach 1: the 2017 protocol}

\begin{table}[t]
\caption{Provenance of the protocol: 2017 main text \cite{dasgupta2017},
2018 FlyLSH code \cite{sharma2018}, or our choice.}
\label{tab:prov}
\centering
\scriptsize
\setlength{\tabcolsep}{3pt}
\begin{tabular}{p{1.7cm}p{3.9cm}p{2.0cm}}
\toprule
element & choice & source \\
\midrule
data & $10{,}000$ vectors: SIFT-small base, GloVe 6B 300d and MNIST train (seeded subsets) & 2017 (size); ours (subset) \\
queries, truth & $1{,}000$ queries; top $2\%$ by Euclidean distance on raw features & 2017 \\
preprocessing & each vector centred by its own mean & 2017 \\
fly & $m=20k$ or $10d$; top-$k$ winners & 2017 \\
sampling & each cell samples $10\%$ of inputs & 2018 code \\
tag & binary; Hamming distance, random tie-break & 2018 code; ours \\
LSH & $k$ real-valued Gaussian projections & 2017 \\
score & AP@200 and recall@200 & ours (2017: undefined ``mAP'') \\
trials & $50$; matrices and queries redrawn & 2017 \\
\bottomrule
\end{tabular}
\end{table}

Table~\ref{tab:prov} gives the protocol and the source of each element; we add
$10{,}000$ mixtures of DoOR odour profiles as a fourth dataset. The 2017 paper
reports ``mean average precision'' without a formula. Precision averaged over
\emph{retrieved} true neighbours only, a convention in parts of the hashing
literature, reproduces its LSH baseline (MNIST, $k=4$: \ReplMnistLshRetr{}
against $0.160$) but not its fly score (\ReplMnistFlyRetr{} against $0.448$).
That score ignores misses, so we report AP@200 and recall@200 and compare with
the paper through orderings and ratios. With a connectome, $M$ accepts only its
own glomeruli: odours use the glomeruli DoOR covers, and SIFT, GloVe and MNIST
are reduced by PCA to one component per glomerulus of that connectome ($49$ or
$51$), assigned to glomeruli by a fresh random permutation in each trial, so these datasets test the wiring's
statistics, not its odour-specific pairing. Each connectome is scored against
$\ConnB$ nulls and one draw of each control per trial.

\subsection{Approach 2: odour analysis}
\label{sec:odours}

Responses are the DoOR 2.0 consensus matrix \cite{munch2016} (commit
\texttt{\DoorCommit}). Coverage is uneven: $35$ glomeruli are measured for $51$
or more odorants, eleven for $12$ to $24$, five for none. We keep the $35$ (our
threshold of $40$ odorants falls in that gap) and odorants measured on at least
$12$ of them: $\NOdorants$ odorants, \MissingPct{} of entries unmeasured and set
to zero. The benchmark is $\NItems$ linear mixtures of two to five profiles,
normalised to unit mean as in the circuit, with the $\kappa=10$ nearest items
on raw responses as ground truth. The primary size $k=\PrimaryK$ ($5\%$ of
cells) and secondary sizes $\{8,16,32,64\}$ were fixed in an earlier revision,
after first results had been seen; this is not a preregistration. MaleCNS is
tested against \NullB{} nulls with a two-stage bootstrap (resampled odorants,
regenerated mixtures, $20$ nulls drawn from the same pool per replicate,
$\BootR$ replicates) and a $\pm\MarginPrimary$ equivalence margin; every
hemisphere is tested against \AnimBOdour{} nulls on its own DoOR glomeruli and
against \AnimB{} on the $\AnimCommonOdour$ shared ones. Null ensembles of this
size, and the audit of weak connections below, were pre-registered
(\texttt{experiments/PREREGISTRATION\_2026-09-28.md}) after an earlier
version of this analysis had used $100$ nulls.

\section{Results}

\subsection{The 2017 pattern holds}

\begin{figure*}[t]
\centerline{\includegraphics[width=\textwidth]{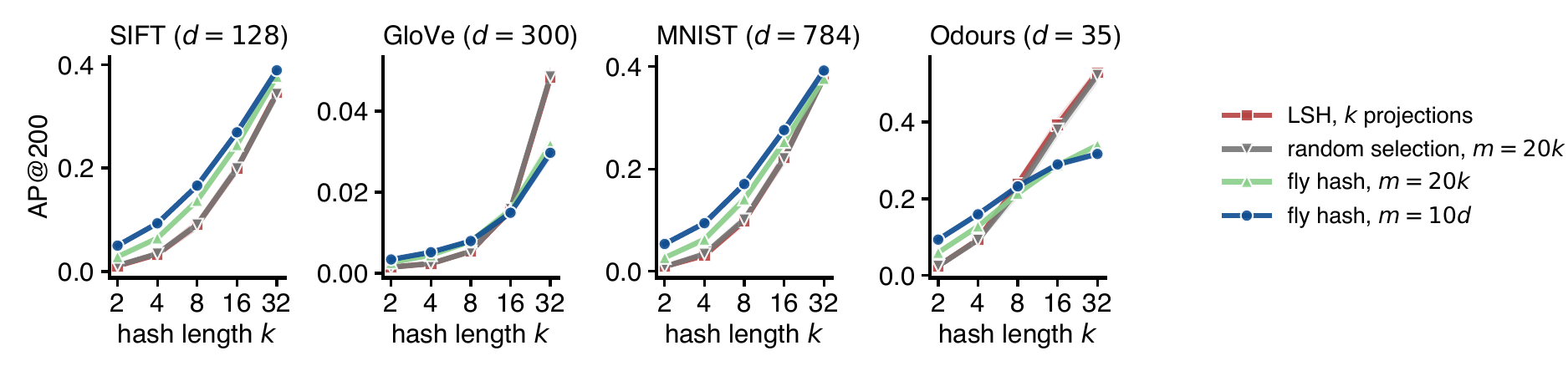}}
\caption{The 2017 protocol with random matrices: AP@200 against hash length,
mean $\pm$ SD of \ReplTrials{} trials. Random tag selection performs like LSH;
winner-take-all at $m=20k$ and $10d$ beats both at short hash lengths.}
\label{fig:repl}
\end{figure*}

With random matrices the published ordering is recovered on SIFT, MNIST and
odours (Fig.~\ref{fig:repl}): random tag selection performs like LSH, as
\cite{dasgupta2017} argue it must, winner-take-all at $m=20k$ beats both, and
$m=10d$ does better still. The advantage is largest at short hash lengths
($\ReplMnistRatio\times$ on MNIST at $k=4$; recall $\ReplMnistRecallRatio\times$;
$\ReplOdourRatio\times$ on odours) and shrinks with $k$; on odours and GloVe
LSH overtakes the fly hash from $k\approx8$--$16$. Scoring against neighbours
on centred input, as the FlyLSH code does, leaves the MNIST ratio at
$\ReplMnistRatioCentred\times$. GloVe is close to chance at small $k$ for every
method.

\subsection{Against real-valued LSH, the advantage is per active cell, not per operation}

\begin{table}[t]
\caption{Odour analysis, MaleCNS: AP@10 of the fly hash and of one-bit
Gaussian codes (bits in parentheses).}
\label{tab:arch}
\centering
\scriptsize
\setlength{\tabcolsep}{3pt}
\begin{tabular}{rcccc}
\toprule
$k$ & fly & sign, $k$ bits & storage-matched & operation-matched \\
\midrule
\ArchRows
\bottomrule
\end{tabular}
\end{table}

\begin{figure*}[t]
\centerline{\includegraphics[width=\textwidth]{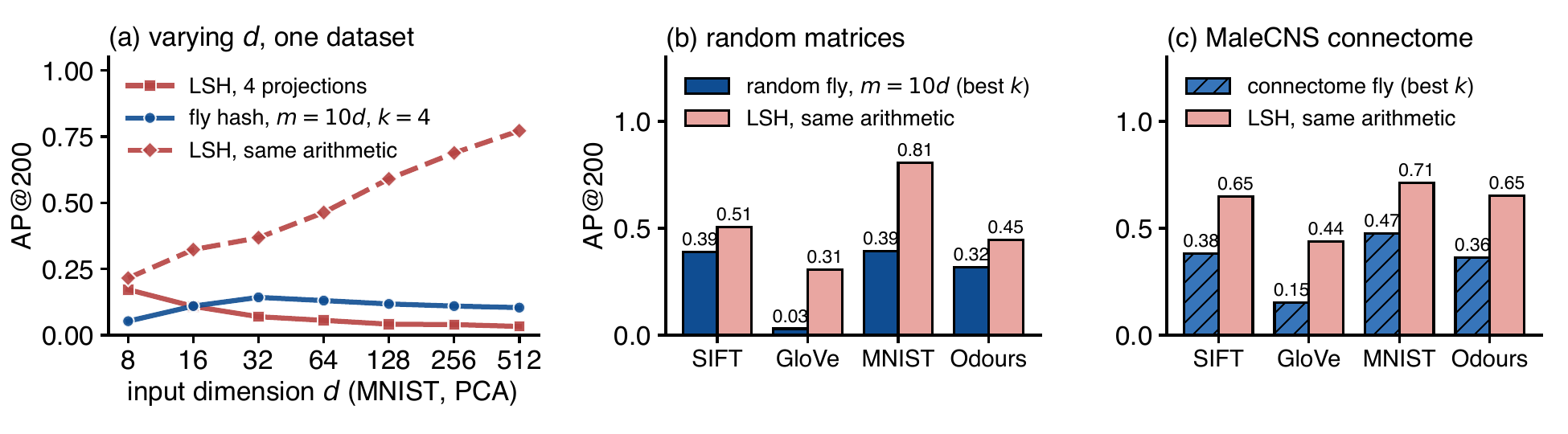}}
\caption{(a) One dataset at several input dimensions (MNIST reduced by PCA to
$d$; PCA also changes the retained information and the integer sampling rate),
$k=4$: the fly hash's lead over LSH per active cell grows with $d$, but LSH
given the same projection arithmetic as the $m=10d$ fly leads at every $d$.
(b, c) Best AP of the fly hash over the tested hash lengths against real-valued
LSH with the same projection arithmetic: (b) random matrices on the original
inputs, (c) the MaleCNS connectome on its own inputs. Inputs and ground truth
differ between (b) and (c); only pairs within a panel are comparable.}
\label{fig:budget}
\end{figure*}

The 2017 comparison sets $k$ active cells against $k$ projections, which
equates projection arithmetic only at $m=20k$. The headline network has
$m=10d$ cells sampling about $d/10$ inputs each: roughly $d^2$ additions
against $2kd$ for LSH, a factor near $d/2k$ ($98\times$ on MNIST at $k=4$). Varying
the input dimension of one dataset (Fig.~\ref{fig:budget}a) shows both sides:
per active cell, the fly hash trails LSH at $d=8$, ties at $16$ and leads from
$32$ on, though not monotonically; given equal arithmetic, real-valued LSH
leads at every $d$. With a random tie priority among Kenyon cells the fly/LSH
ratio at $k=4$ is \TieRatioEight{} at $d=8$ and \TieRatioSixteen{} at $d=16$
(index order: \IdxRatioEight{} and \IdxRatioSixteen{}): the fly hash trails at
$d=8$ and reaches parity by $d=16$ under both rules, but the value at $d=8$
depends on the rule. The same holds on all
four datasets (Fig.~\ref{fig:budget}b, c), for random matrices (LSH
\ReplOpsMin{}--\ReplOpsMax{} AP) and for the connectome, whose
$\mathrm{nnz}(M)$ additions buy $\ConnOpsBits$ projections on image and word
input and $\ConnOdourOpsBits$ on odours; these beat it on \ConnOpsWins{} of
\ConnNData{} datasets. ``Best'' is the best over the tested hash lengths.

One-bit codes are weaker than real-valued projections, and against them the
comparison is closer (Table~\ref{tab:arch}): on odours the fly hash beats sign
codes of $k$ bits ($\ArchKbitsRatio\times$ at the primary size), trails at
matched storage ($\ArchStorageRatio\times$), and at matched operations trails
for small $k$ and leads from $k=32$ ($\ArchComputeRatio\times$).

\subsection{Connectomes behave like degree-preserving random matrices}

\begin{table}[t]
\caption{Odour analysis, MaleCNS: connectome against the null ($B=\NullB$) and
comparison matrices (relative to the null mean, \%). $^\ast$primary.}
\label{tab:wiring}
\centering
\scriptsize
\setlength{\tabcolsep}{2.4pt}
\begin{tabular}{rcccccccc}
\toprule
$k$ & conn. & null & $\Delta$\% & $p$ & $p_\text{Holm}$ & fan flat & 6 unif. & balanced \\
\midrule
\WiringRows
\bottomrule
\end{tabular}
\end{table}

\begin{table}[t]
\caption{Odour analysis, MaleCNS, primary size, one-factor variations ($B=\RobB$
nulls each): relative difference (\%) and two-sided $p$.}
\label{tab:rob}
\centering
\scriptsize
\setlength{\tabcolsep}{3pt}
\begin{tabular}{lcc}
\toprule
variation & $\Delta$\% & $p$ \\
\midrule
\RobRows
\bottomrule
\end{tabular}
\end{table}

\begin{figure*}[t]
\centerline{\includegraphics[width=\textwidth]{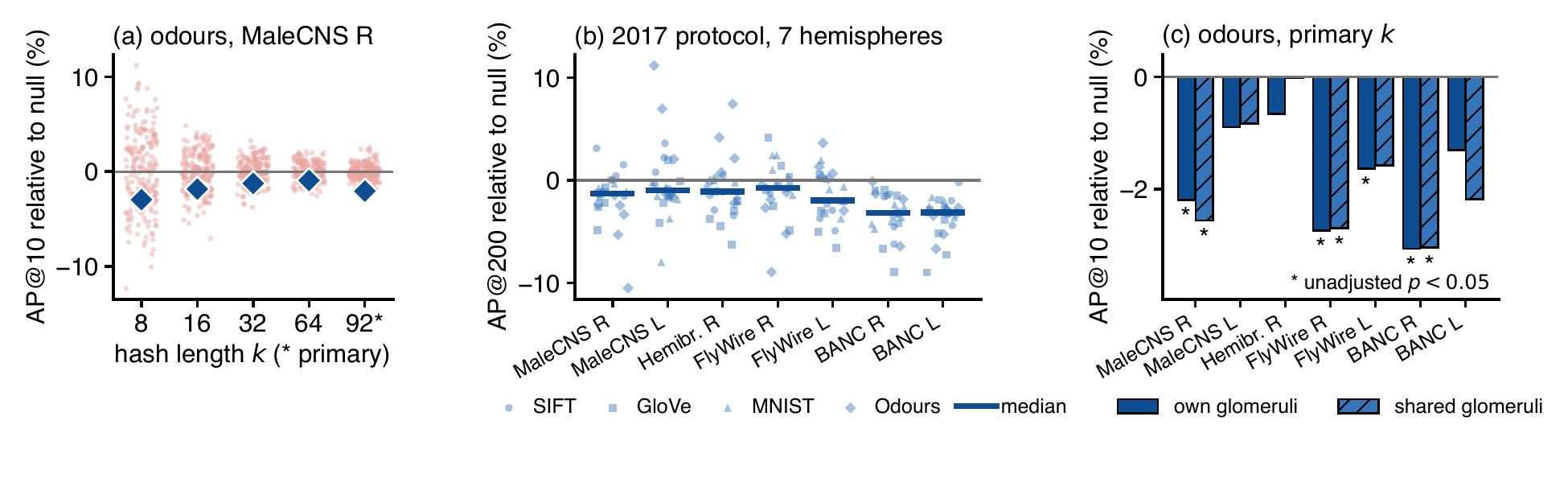}}
\caption{Connectome against its own null (AP relative to the null mean). (a)
Odour analysis, MaleCNS right: orange, \NullB{} null matrices; blue, the
connectome. (b) The 2017 protocol in seven hemispheres: four datasets $\times$
six hash lengths each, bar at the median. (c) Odour analysis at the primary
size in every hemisphere, on its own DoOR glomeruli and on the
$\AnimCommonOdour$ all share.}
\label{fig:conn}
\end{figure*}

\begin{table*}[t]
\caption{Seven hemispheres of four animals: Kenyon cells, mean inputs per
cell, fan-out coefficient of variation, co-occurrence $Q$ as $z$ against
$\AnimBQ$ nulls; odour analysis at the primary size on own glomeruli
($\AnimBOdour$ nulls, floor $\AnimOdourPFloor$) and on shared glomeruli
($\AnimB$ nulls) (relative difference \%, unadjusted $p$);
median over the 2017-protocol runs ($4$ datasets $\times$ $6$ sizes, $B=\AnimBBench$,
\AnimTrials{} trials); fly hash over LSH on MNIST at $k=4$.}
\label{tab:animals}
\centering
\scriptsize
\setlength{\tabcolsep}{5pt}
\footnotesize
\begin{tabular}{lccccccccc}
\toprule
hemisphere & sex & cells & inputs & fan CV & $Q$ $z$ & odours $\Delta$\% ($p$) & shared $\Delta$\% ($p$) & protocol median $\Delta$\% & fly/LSH \\
\midrule
\AnimalRows
\bottomrule
\end{tabular}
\end{table*}

Substituting a connectome for the random matrix keeps the fly hash's advantage:
it leads LSH on MNIST at $k=4$ by \AnimLshMin{}--\AnimLshMax$\times$ in every
hemisphere (Table~\ref{tab:animals}). Against its own null it is close but
slightly worse (Fig.~\ref{fig:conn}b): across $\AnimNProto$ hemisphere,
dataset and size combinations the median difference is \AnimRelMedian{}
(interquartile range \AnimRelIQR), negative in \AnimNNeg{}, below every null
draw in \AnimNFloor{} and above every draw in \AnimNFloorUp{}, with a negative
mean in \AnimalsProtoNeg{} of \AnimAnimals{} animals. These are descriptive
counts over dependent conditions, not independent replications. Seven hemispheres come
from four animals and are not independent; one male cannot separate sex from
individual or reconstruction differences.

On odours (Fig.~\ref{fig:conn}c) the difference at the primary size is
negative in all \AnimN{} hemispheres, \AnimOdourMin{} to \AnimOdourMax{}, and
\AnimMatchedMin{} to \AnimMatchedMax{} on the shared glomeruli. The sign is
consistent: \AnimOdourNSig{} unadjusted $p$-values are below $0.05$, and after Holm
adjustment across the seven hemispheres \AnimOdourHolmN{} remain, \AnimOdourHolmNames{}
(smallest adjusted $p=\AnimOdourHolmMin$; on the shared glomeruli, with the
smaller ensemble, the smallest is $\AnimMatchedHolmMin$). In the full MaleCNS analysis (Fig.~\ref{fig:conn}a, Table~\ref{tab:wiring}) the connectome scores $\PrimaryReal$ against
$\PrimaryNullMean\pm\PrimaryNullSD$, \PrimaryRel{} (two-sided
$p=\PrimaryPtwo$; $\PrimaryPHolmAll$ with all five sizes as one family); the
bootstrap $90\%$ interval is \BootLo{} to \BootHi{}, which lies within the
chosen $\pm\MarginPrimary$ margin but not within $\pm2.5\%$; in a simulation
from one fitted generator, this interval covered the true value in \CovNew{}
of \CovDatasets{} datasets, which calibrates it for that generator only, and the two-sided test has $80\%$ power at
\MdeDown{} downward and \MdeUp{} upward. Across \RobN{} one-factor variations (Table~\ref{tab:rob}) the deficit ranges from
\RobRelMin{} to \RobRelMax{}, and it depends on the input: imputing unmeasured responses gives \ImputeRelMin{} to
\ImputeRelMax{} ($p\ge\ImputePMin$), and all $46$ glomeruli with data give no
difference.

By $Q$ the pairing departs from its null in \AnimQNSig{} of \AnimN{} hemispheres
($Q$ $z$ from \AnimQzMin{} to \AnimQzMax{}; \AnimQNSigHolm{} after Holm adjustment
over seven, adjusted $p\le\AnimQHolmMax$), consistent with \cite{zheng2022}.
The hemibrain is indistinguishable from its null when every connection is
counted, but not once single-synapse connections are removed: at two or more
synapses per glomerulus--cell pair its $Q$ is above the null
($z=\HemiQzTwo$, $p=\HemiQpTwo$), as it is in all \QDepartTwo{} hemispheres. Its
apparent exception is a property of its weakest connections. Under these protocols the observed structure did not coincide with
a consistent retrieval advantage; seven hemispheres do not allow a general
statement about structure and retrieval.

\subsection{Equalising fan-out improves retrieval in the model}

\begin{figure*}[t]
\centerline{\includegraphics[width=\textwidth]{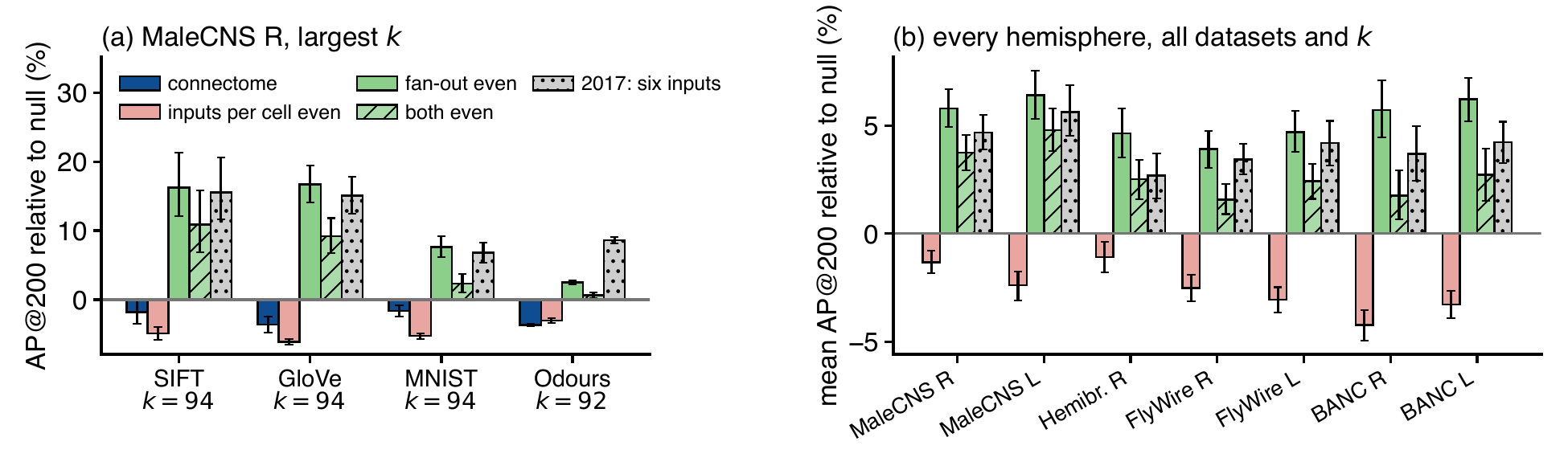}}
\caption{Equal-connection degree controls, AP@200 relative to the null mean
(ratio of trial means), with $95\%$ bootstrap intervals over trials, each trial
paired with its own null mean. (a) MaleCNS right at the largest hash size
(\CtrlTrials{} trials, $\CtrlB$ nulls). (b) All hemispheres, averaged over the
four datasets and six hash sizes.}
\label{fig:ctrl}
\end{figure*}

The comparison with the null concerns the pairing alone: connectome and null
share both degree sequences, so neither sequence can explain the difference
between them. We asked separately whether the measured degree sequences limit
retrieval relative to more regular random matrices at the same connection count,
changing one sequence at a time (Fig.~\ref{fig:ctrl}). These are interventions
within the model, compared with each trial's own null mean.

In MaleCNS right, making fan-out even, so that every glomerulus reaches the
same number of Kenyon cells, raises AP on image and word input by \CtrlOutMin{}
to \CtrlOutMax{} across datasets and hash sizes, with the interval above zero in
\CtrlOutPos{} of \CtrlCells{} cases, and on odours by \CtrlOutOdMin{} to
\CtrlOutOdMax{} (\CtrlOutOdPos{} of \CtrlOdCells{}). Making inputs per cell even
has mixed effects: \CtrlInMin{} to \CtrlInMax{} on image and word input (interval
above zero in \CtrlInPos{}, below in \CtrlInNeg{}), and \CtrlInOdMin{} to
\CtrlInOdMax{} on odours, largest at the smallest hash sizes. Averaged over
datasets and sizes (Fig.~\ref{fig:ctrl}b), even fan-out raises AP in all
\AnimCtrlN{} hemispheres (\AnimOutLo{} to \AnimOutHi{}, every interval above zero),
whereas even inputs per cell lowers it in all \AnimCtrlN{} (\AnimInLo{} to
\AnimInHi{}, every interval below zero); its gains are confined to some small
hash sizes. The six-input
construction of 2017 behaves like the even-fan-out control. The measured
fan-out is strongly skewed (up to $\FanMax$ cells for \FanMaxGlom{} in MaleCNS right, against a mean near
$195$). This shows that more even fan-out would
retrieve better at equal cost in this model, not that the skew is a biological
cost: it may serve other ends.

Row-centring, part of the 2017 protocol, alters winner order only when inputs
per cell differ. Removing it changes the connectome's AP on image and word input
by \NoCentSmallMin{} to \NoCentSmallMax{} at the smallest hash size and by at most
\NoCentLargeAbs{} at the largest, and changes the odour results strongly at small
$k$, where drive on non-negative input then tracks the number of inputs
(supplement).

\subsection{Novelty detection and nearest-neighbour classification}

\begin{table}[t]
\caption{Novelty detection and nearest-neighbour classification (MaleCNS
right, \TaskTrials{} trials): null mean, and difference to it for the
connectome, the equal-connection controls and the six-input construction.
$^\dagger$95\% interval excludes zero.}
\label{tab:tasks}
\centering
\scriptsize
\setlength{\tabcolsep}{2.5pt}
\begin{tabular}{lcccccc}
\toprule
task & null & conn. & fan-out even & inputs even & both & six \\
\midrule
\TaskRows
\bottomrule
\end{tabular}
\end{table}

The fly hash has also been proposed as a novelty detector, a Bloom filter whose Kenyon-cell output synapses are depressed by stored items \cite{dasgupta2018}, and as a nearest-neighbour classifier with one such filter per class \cite{ram2022}. We use a static graded surrogate of that filter, not a reproduction of its dynamics: after storing $n$ items, cell $j$'s output weight is $w_j = 1 - c_j/n$ with $c_j$ the number of stored items that activated it, and a query's novelty is the mean $w_j$ over its $k$ active cells (the binary filter saturates at these sizes); the classifier predicts the class whose filter finds the query least novel. This analysis is exploratory: MaleCNS right, \TaskTrials{} trials, $10$ null matrices reused across trials, unadjusted $95\%$ intervals of the difference to each trial's null mean. Protocol: MNIST uses $6{,}000$ images per trial, PCA to one component per glomerulus with a random component-to-glomerulus assignment, row-centred, $4{,}000$ for storing and $2{,}000$ held out as queries, $k=\TaskKMnist$ ($5\%$ of the cells); odours use $3{,}000$ stored mixtures and $500$ familiar plus $500$ novel queries on the measured glomeruli, $k=\TaskKOdour$. With these settings,
the connectome differs from its null by \TaskNovConn{} in MNIST novelty AUC
(null \TaskNovNull; digits 0--4 stored, held-out 0--4 familiar, 5--9 novel), \TaskKnnConn{} in ten-class accuracy (null \TaskKnnNull) and \TaskOdNovConn{} in odour novelty AUC (null \TaskOdNovNull; mixtures of half of the odorants stored, mixtures of the other half novel). Even fan-out gives \TaskNovOut, \TaskKnnOut{} and \TaskOdNovOut;
even inputs \TaskNovIn, \TaskKnnIn{} and \TaskOdNovIn{} (Table~\ref{tab:tasks}). The pairing again gives no advantage. Even fan-out raises MNIST classification accuracy, has no resolvable effect on MNIST novelty, and lowers odour novelty in this experiment.

\subsection{The fan-out profile is similar across the sampled animals}

Across the four sampled connectomes the fan-out profile is similar
(Fig.~\ref{fig:concept}b): over the $\FanCommon$ glomeruli all hemispheres
share, relative fan-out correlates between animals at Spearman
$\rho=\FanBetweenMin$ to $\FanBetweenMax$ (median $\FanBetweenMed$) and between the
two hemispheres of one animal at $\FanWithinMin$ to $\FanWithinMax$; it spans a
factor of about $\FanSpread$, from \FanTop{} at the top to \FanBottom{} at the
bottom. This is a descriptive comparison of four animals, not a population
estimate, and shared cell-type definitions, nomenclature mapping or systematic
reconstruction biases could contribute to the similarity. That some glomeruli
are over-represented was reported by \cite{caron2013} and, as structured
sampling of food-related types, by \cite{zheng2022}; new here is the comparison
across reconstructions and its retrieval cost in the model. A separate
anatomical observation is that structural synapse counts do not offset the
skew: the mean number of synapses per connection correlates positively with
fan-out in \FanWeightPos{} of \FanN{} hemispheres ($\rho$ from $\FanWeightMin$ to
$\FanWeightMax$, nominal $p<0.05$ in \FanWeightSig, unadjusted), so in those hemispheres the summed synapse count per glomerulus is even more uneven than fan-out. These are counts of reconstructed synapses, not physiological efficacies, and the retrieval model is binary and does not use them; the retrieval cost above was shown for binary fan-out only. Its relation to odour
tuning in DoOR is weak: over measured odorants, fan-out correlates with
response breadth at $\rho=\FanBreadthMin$ to $\FanBreadthMax$ (smallest
$p=\FanBreadthPmin$) and with response SD at $\FanSdMin$ to $\FanSdMax$ (nominal
$p<0.05$ in \FanSdSig). These seven dependent correlations per measure share one odour panel and
similar fan-out profiles; we treat them as descriptive, uncorrected
associations rather than independent replications.

\section{Discussion}

Under these retrieval protocols, the measured pairing gives no consistent
advantage over degree-preserving random wiring: four connectomes keep the fly
hash's per-cell advantage and retrieve within a few percent of their rewirings,
slightly lower, in the direction reported for odour discrimination by
\cite{zheng2022}; the small odour deficits depend on the input model. The
2017 construction, which differs from both in its degree sequences, retrieves
better than either, and the controls attribute that mainly to its even fan-out.
The measured skew in fan-out may serve other ends, such as innate valence
\cite{zheng2022} or novelty detection \cite{dasgupta2018}, which we do not
test.

The algorithmic claim needs a qualifier. Against the tested real-valued Gaussian
baseline the fly hash's advantage is per active cell: Gaussian projections given the same projection arithmetic
retrieve better on every dataset and input dimension we tested. Against one-bit
codes, and at the circuit's own scale, the comparison is closer. The random construction of 2017 already has nearly even fan-out, so the
fan-out result is not an improvement over the standard fly hash; it locates
where the circuit departs from it. That departure looks similar in all four sampled animals, and structural synapse counts do not offset it. In the binary model, a circuit that gives more Kenyon cells to the same glomeruli pays for it in generic similarity search under these protocols (that most sampled hemispheres also place more synapses per connection on those glomeruli is an anatomical observation the model does not use); what it gains in return, for instance in innate or learned valence of the food-related glomeruli near the top of the ranking \cite{zheng2022}, is a question these results raise but do not answer.

\section{Limitations}

Four animals, one male; differences between connectomes mix biology and
reconstruction. Our reimplementation recovers the published pattern but not
its absolute values, and two protocol details come from later code. Image and
word input is compressed to one PCA component per glomerulus and assigned to
glomeruli arbitrarily. The cross-connectome protocol runs use $\AnimBBench$ nulls and are
effect estimates. The odour deficit depends on treating unmeasured DoOR entries
as zero. The model is binary and rate-free and ignores inhibition beyond
winner-take-all.

\section{Conclusion}

In a documented reimplementation, sparse winner-take-all codes outperform
equally many real Gaussian projections at short hash lengths. Across four
connectomes, measured wiring gives no consistent retrieval advantage over
degree-preserving random wiring, and is slightly worse on average under the tested inputs. Separately, in the model, the circuit's skewed glomerular fan-out, which is similar across the sampled animals, costs retrieval relative to even fan-out at equal connection count. A fly hash
therefore needs no connectome data, and in the tested generic-retrieval model
random sparse wiring with even fan-out does at least as well as the measured
wiring. The algorithm's advantage over real-valued LSH is per active unit, not
per operation: where arithmetic is cheap, dense Gaussian projections retrieve
better; where active units are the binding cost, as may hold for sparse or
neuromorphic hardware, the fly hash could be the better choice, which direct
hardware measurement would have to show. The pre-registration that fixed the
null ensembles of Section~\ref{sec:odours} also contained three algorithmic and
theory experiments outside this comparison; none met its registered criterion, and \texttt{experiments/OUTCOMES\_2026-09-28.md} reports them.
Code, data checksums, results and a supplement are at
\url{https://github.com/ssenge/FlyHash-Connectome}.

\section*{Acknowledgment}

Claude (by Anthropic) was used as a coding and analysis assistant; the author
designed the study.

\end{document}